\documentclass[11pt]{article}

\usepackage[preprint]{acl}

\usepackage{times}
\usepackage{latexsym}

\usepackage[T1]{fontenc}

\usepackage[utf8]{inputenc}

\usepackage{microtype}

\usepackage{inconsolata}

\usepackage{amsmath}
\usepackage{booktabs}
\usepackage{todonotes}
\usepackage{xcolor}
\usepackage{graphicx}
\usepackage{multirow}
\usepackage{makecell}
\usepackage{subcaption}
\usepackage{amsmath}
\usepackage{amssymb}
\usepackage{langsci-gb4e}
\usepackage{tabularx}
\usepackage{tikz}
\usepackage{enumitem}
\usepackage[most]{tcolorbox}
\usepackage{subcaption}

\newcommand*\circled[1]{\tikz[baseline=(char.base)]{
            \node[shape=circle,draw,inner sep=.6pt] (char) {#1};}}

\definecolor{paprikaRed}{RGB}{170, 36, 36}
\definecolor{danubeBlue}{RGB}{40, 84, 135}
\definecolor{tokajiGold}{RGB}{218, 165, 32}
\definecolor{matyoGreen}{RGB}{32, 102, 61}

\title{When Discourse Pressures Conflict: \\ Information Structure in Vision-Language Model Outputs}

\author{Marcell Fekete$^{1}$ \ \ Johannes Bjerva$^{1}$ \ \  Tamás Káldi$^{2,3}$ \\
         $^1$Department of Computer Science, Aalborg University, Copenhagen, Denmark; \\
          $^2$Department of Psycholinguistics and Neurolinguistics, \\ ELTE Research Centre for Linguistics, Budapest, Hungary \\
          $^3$ELTE Bárczi Gusztáv Faculty of Special Needs Education, Budapest, Hungary \\
          $^1$\texttt{\{mrfe, jbjerva\}@cs.aau.dk}; $^2$\texttt{kaldi.tamas@nytud.hu} \\ 
}

\begin{document}
\maketitle
\begin{abstract}
Vision-language models (VLMs) are increasingly evaluated for whether they identify the right visual content, but little is known about whether they express such content in a discourse-appropriate form.
We address this research gap using \textbf{information structure} (\textsc{IS}), testing whether VLMs distinguish discourse-old Topics from discourse-new Foci in visually grounded question answering.
We exploit Hungarian, a language in which Topic and Focus map onto dedicated syntactic positions, making \textsc{IS} choices observable in text.
Comparing six VLMs with human participants, we find that models produce \textsc{IS}-relevant constructions, but over-regularise this sensitivity.
Under the interacting pressures of discourse status, grammatical role (preference for subject Topics) and definiteness (preference for indefinite Foci), humans choose variable strategies for \textsc{IS} realisation.
VLMs, by contrast, collapse onto narrow response templates, resembling mode collapse \citep{kirk2024understanding}.
Our findings suggest that VLM evaluation should look beyond content accuracy to how content is packaged for the discourse.
\end{abstract}

\section{Introduction}\label{sec:intro}

Vision-language models (VLMs) are increasingly used as general-purpose systems for generating language grounded in visual inputs.
Their language abilities, crucial to their downstream capabilities, are typically evaluated through linguistically simple tasks that ask whether a model identifies objects, relations, actions, or answers questions about an image correctly.
Such evaluations are necessary for general benchmarking, but leave open a specifically linguistic research question: when a model has identified relevant content, does it express that content in a form that is appropriate to the discourse context?

This question matters because natural language generation is more than simply a matter of selecting the right content.
Speakers of natural languages package information according to what is already established in the discourse, and what is newly contributed.
Such packaging of information is realized via information structure (\textsc{IS}).
IS has different components, such as the Topic that serves to link the utterance to discourse-old material, and the Focus, which conveys discourse-new or contrastive information \citep{reinhart1981pragmatics,krifka2008basic}.
\textsc{IS} is well-studied in theoretical linguistics, psycholinguistics, and discourse processing  yet it remains under-represented in the evaluation of contemporary LLMs and VLMs.
We know little about whether VLMs can use linguistic form to distinguish between what is already under discussion, from what is newly asserted.

\begin{figure}[t]
    \centering
    \includegraphics[width=\linewidth]{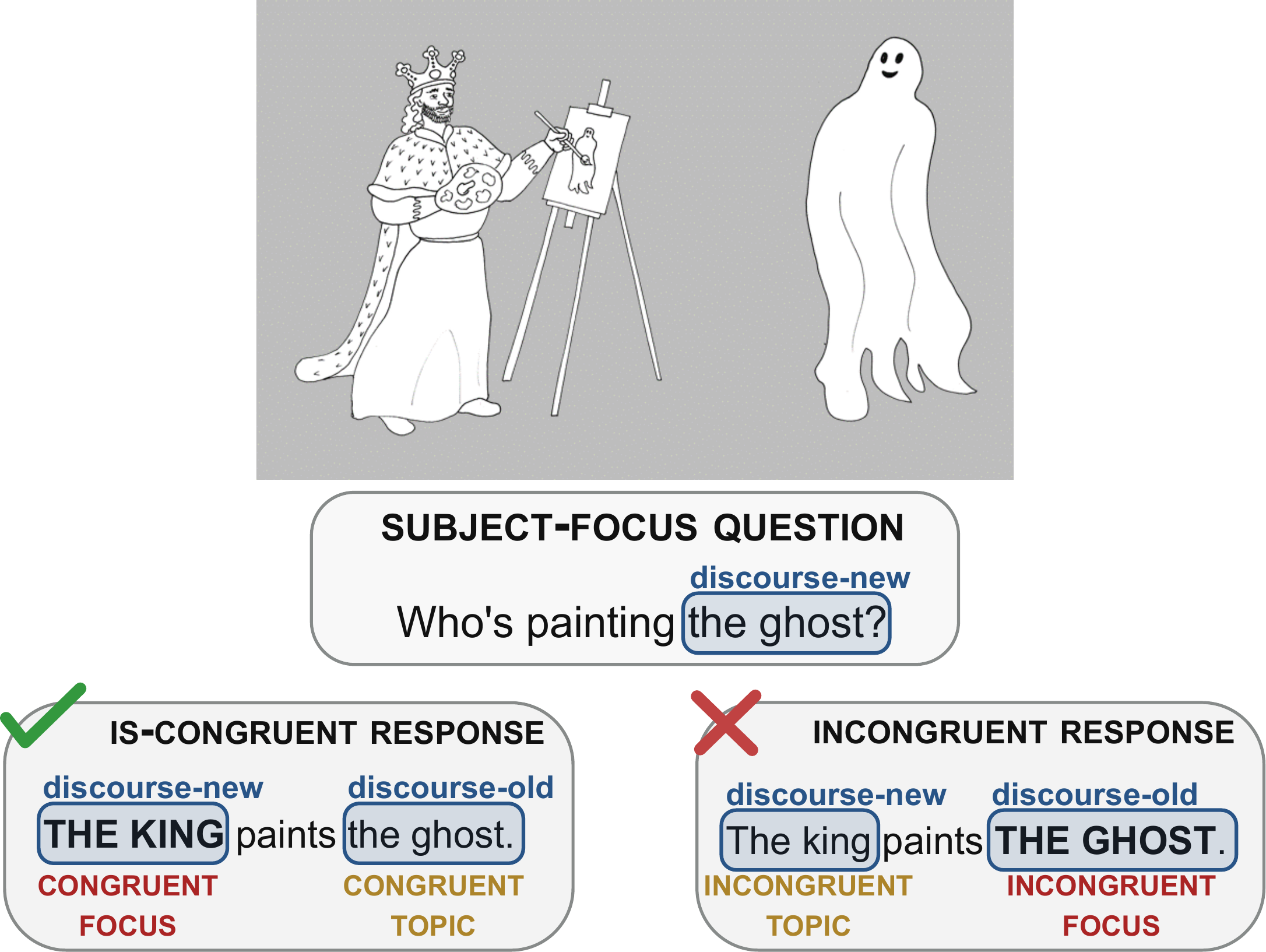}
    \caption{
    Information structure describes how new information is established (Focus) and how already established information is managed (Topic).
    In our paradigm, a wh-question establishes the object (\textit{the ghost}) as discourse-old and elicits the subject (\textit{the king}) as discourse-new.
    An \textsc{IS}-congruent response marks the discourse-new referent as Focus and the discourse-old one as Topic; an incongruent response reverses this assignment.
    English marks Focus prosodically (shown here with capitalisation); Hungarian marks it through word order, making \textsc{IS} choices observable in text.
    }
    \label{fig:figure_1}
\end{figure}

\textsc{IS} is an important phenomenon and it can speed up how humans process utterances \citep{dahan_accent_2002}, while also impacting the truth value of assertions.
For instance, "English is spoken in \textsc{The Shetlands}" -- contrastive emphasis on \textsc{The Shetlands} -- falsely implies that the language is spoken exclusively there and nowhere else \citep{hajicova_plea_2019}.
However, it is difficult to evaluate \textsc{IS} in English text alone, since there it is typically realised prosodically, and is therefore not directly visible in written responses.
We therefore turn to Hungarian, a \textit{discourse-configurational} language in which Topic and Focus are associated with dedicated structural positions \citep{ekiss1995discourse,ekiss2002syntax}.
This makes \textsc{IS} choices observable in text, and Hungarian an exceptionally well-suited testbed for asking whether VLMs produce  discourse-congruent responses.

We study this question using a visual question answering paradigm adapted from psycholinguistic work.
Human participants and VLMs see an image depicting a transitive event, and answer a question that establishes either the subject or the object as discourse-old.
The appropriate response must identify the visually grounded event, but it also has to produce a response with the appropriate \textsc{IS}: to arrange Topic and Focus correctly.
This allows us to test whether VLMs appropriately use Hungarian word order to mark discourse-old and discourse-new referents with appropriate \textsc{IS} positions (see Figure \ref{fig:figure_1}).
It also allows us to examine how this behaviour interacts with two independent pressures known to affect human language production: the tendency for subjects to be topical \citep{givon1983topic,lambrecht1994}, and the association between definiteness and discourse status \citep{hankamer1971constraints,givon1983topic}.

We ask the following research questions:

\begin{enumerate}%

    \item \textbf{RQ1:} Do VLMs exhibit sensitivity to \textsc{IS} when describing images?
    Specifically, do they use syntactically dedicated Topic and Focus positions to distinguish discourse-old and discourse-new elements in response to questions with differing focus, similarly to human speakers?
    
    \item \textbf{RQ2:} How does core \textsc{IS} sensitivity interact with additional pressures known to interact with \textsc{IS} realisation in human production?

    \begin{enumerate}%
    
        \item To what degree does the subject-object asymmetry in VLM topicalisation mirror the well-documented human preference to topicalise subjects?

        \item Do Topics tend to be definite and Foci indefinite, as predicted by discourse-theoretical accounts (definite Topics, indefinite Foci)?
        
    \end{enumerate}
    
\end{enumerate}

\noindent
Our contributions are the following: 
\circled{1} The \textbf{first systematic investigation of \textsc{IS} realisation in VLM outputs}, addressing a blind spot in VLM evaluation. 
\circled{2} A \textbf{linguistically controlled multimodal and interdisciplinary evaluation framework} that adapts a psycholinguistic paradigm for VLMs, using Hungarian to make Topic--Focus choices observable in text.
\circled{3} Empirical evidence that while humans show variable strategies in order to realise \textsc{IS}, \textbf{VLMs partially track discourse context, but over-regularise interacting linguistic pressures and converge on narrow response templates} that resemble mode collapse.

\section{Background}\label{sec:background}

\subsection{Linguistic Evaluation of VLMs}

When it comes to evaluating the core linguistic abilities of language models, emphasis falls on syntax and sentence-level semantic phenomena across a range of phenomena, including subject-verb agreement, anaphor agreement, long-distance dependencies, and NPI licensing \citep{warstadt-etal-2020-blimp,someya-oseki-2023-jblimp,taktasheva-etal-2024-rublimp,jumelet-etal-2026-multiblimp}.
Certain pragmatic phenomena have received growing attention with dedicated benchmarks focusing on implicature, presupposition, and deixis \citep{ma-etal-2025-pragmatics}.
Discourse-level phenomena are addressed but more often only indirectly, through downstream tasks such as summarisation, document-level translation, and long-context question-answering \citep{ma-etal-2025-pragmatics}.
Despite its importance for human language processing \citep{dahan_accent_2002,hajicova_plea_2019}, \textsc{IS} is essentially absent from this landscape, with the partial exception of \citet{cuneo-etal-2025-gpt4}, a probing study targetting a specific \textsc{IS}-syntax relationship in English rather than a general evaluation framework.
To our knowledge, no benchmark or evaluation framework dedicated to \textsc{IS} exists for either LLMs or VLMs, despite its key role in communication.

\subsection{Information Structure (\textsc{IS})}\label{subsec:is}

\textsc{IS} is the partitioning of utterance content according to discourse status: selectively foregrounding and backgrounding elements relative to the common ground built by the interlocutors.
\textsc{IS} reflects speaker intent in guiding the hearer's attention, and as such, it is a communicative phenomenon humans are particularly sensitive to 
\citep{fowler1987talkers,roettger_mapping_2019}.
Its principal primitives are Topic and Focus, typically realised as noun phrases (NPs).
The Topic provides the theme of an utterance, defining what it is \textit{about}, and denotes an already familiar entity in the discourse \citep[p. 118]{lambrecht1994}.
The Focus, on the other hand, conveys new information pertinent to the Topic \citep{reinhart1981pragmatics,heim1982semantics,krifka2008basic,erteschikshir2013information}.
Its function ranges from merely introducing new content (\textit{information focus}) to marking contrast or expressing exhaustive identification over a contextually given set (\textit{identificational focus}) \citep{ekiss1998}.

For instance, in response to the question \textit{Who did Thea kiss?}, the canonical answer is:

\ea\label{ex:thea_canonical}
    $\big[_T$ Thea $\big]$ kissed $\big[_F$ Fred $\big]$.
\z

\noindent
Here `Thea' is the Topic, linking the response to the question, and `Fred' is the (information) Focus, introducing the new referent.
The Topic is optional -- a bare `Fred' is also felicitous -- but the Focus is obligatory \citep[p. 207]{lambrecht1994}, as it is otherwise unrecoverable from context.
\textsc{IS} can also be expressed through periphrastic constructions such as clefting (\textit{It was Fred that Thea kissed}), but its primary exponent in English is prosody, particularly intonation \citep{hajicova_automatic_1995,arnold_information_2013}.

This makes \textsc{IS} an elusive phenomenon to observe in the English-dominated web corpora that serve as the primary training data of LLMs.
Cross-linguistically, however, \textsc{IS} has alternative exponents, some of which are overt in the written medium.
In discourse-configurational languages such as Hungarian, \textsc{IS} is in fact the primary factor shaping word order \citep{ekiss1995discourse,ekiss2002syntax}.
Accordingly, \textsc{IS} functions are assigned dedicated syntactic positions.
The Topic NP occupies the first sentence position, while, in its canonical form, the Focus NP directly precedes the main verb, displacing any other linguistic elements, including verb modifier -- often aspectual or directional -- that is otherwise prefixed to the verb (\textit{igekötő}, glossed as \textbf{I}; see Table \ref{tab:focus_types} for examples).
Optionally, the Focus NP may also appear post-verbally without displacing the verb modifier \citep{ekiss1998,kenesei2006focus}, but this reflects a less marked form of Focus.
The primacy of the preverbal Focus NP makes the displacement of the verb modifier the most straightforward diagnostic of Focus in written Hungarian.
The structural encoding of \textsc{IS} in Hungarian makes the language an ideal testbed for investigating the \textsc{IS} capabilities of VLMs.

\subsection{Pressures on \textsc{IS}}\label{subsec:pressures}

The realisation of \textsc{IS} is shaped by pressures beyond the discourse status itself, ranging from processing constraints such as working memory and attention \citep{sanford2006shallow,sanford2009enhancement,ward2007linguistic,kaldi2020contextual,kaldi2021structural,kaldi2021linguistic} to properties of the linguistic elements themselves \citep{wasow1997remarks,demszky2021role}.
In the following, we introduce two such factors that interact with \textsc{IS} in our experimental setup: grammatical role (\textbf{RQ2a}) and definiteness (\textbf{RQ2b}).

\paragraph{Grammatical role}

A robust cross-linguistic tendency is for grammatical subjects to also function as Topics on the discourse-level, despite the two belonging to separate dimensions of linguistic analysis.
Narrative discourse is typically centred on human or animate referents, who tend to be agents of transitive verbs and are therefore prime candidates for both subjecthood and topicalisation \citep{givon1983topic,aissen1999markedness,haig2016discourse}.
Discourse-configurational languages such as Hungarian show a weaker association of \textsc{IS} and grammatical roles than languages like English, but Hungarian nevertheless favours subject Topics and, more generally, subject-before-object orderings \citep{macwhinney1988processing}.
All else being equal, we therefore expect a bias towards subject Topics in our setup, unless overridden by other cues \citep{lambrecht1994}.

\paragraph{Definiteness}

Beyond the \textsc{IS} functions of Topic and Focus, `givenness' also underlies the distinction between definite NPs -- referring to entities already established in the discourse -- and indefinite NPs, which typically introduce fresh referents into the context \citep{hankamer1971constraints,kuno1972functional,heim1982semantics}.
The same discourse pressures that align subjects with Topics tend to align them with definiteness as well: subjects in discourse are overwhelmingly definite \citep{givon1983topic}.
In Hungarian, definiteness has accordingly been proposed as a predictor of the syntactic position of an NP \citep{ekiss2002syntax}.
A corpus study by \citet{demszky2021role}, drawing on the Hungarian Gigaword corpus \citep{oravecz2014hungarian}, reports only a slight effect, but this likely underestimates the true association.
Corpus data conflate the two Focus types (§\ref{subsec:is}), whereas the definiteness--Focus correlation is theoretically predicted to hold most strongly under identification \citep{ekiss1998}.
In communicative situations where Focus is restricted to its identifying function -- as in our experimental setup -- we therefore expect definiteness to pattern more closely with \textsc{IS} position.

\section{Methodology}\label{sec:methods}

\paragraph{Stimuli and design}

We collected human data from Hungarian speakers to evaluate how accurately VLMs express \textsc{IS}.
Hungarian is uniquely suited for this investigation, as the \textsc{IS} functions of verbal arguments can be directly inferred from word order and the syntactic position of the verb modifier.
Participants -- humans or VLMs -- are shown one of 47 black-and-white cartoon-style images depicting a transitive action between two characters.
Each image is accompanied by one of two \textit{wh}-questions that name one character explicitly and elicit the other: in the \textit{object-focus} condition, the subject is named and the object is queried (see Figure~\ref{fig:figure_1}).
In the \textit{subject-focus} condition, this is reversed.
The named character is thereby established as discourse-old (a candidate Topic in the response), while the elicited character is new (the Focus).
Stimuli were chosen to control for several pressures on \textsc{IS} discussed in §\ref{subsec:pressures}: the events are semantically reversible by making all characters animate and human, removing animacy as a confound; actions are picked to avoid fixed expressions with lexicalised complements (\textit{e.g.}, `walking the dog'); and all verbal expressions contain a verb modifier, so that Focus placement is overtly marked in writing.\footnote{The images are drawn from \citet{kaldi-lukacs-inprep}.}

All the vocabulary items used to name characters are common words, their lemmas have uniformly high frequency counts when counted on a 1B Hungarian corpus (Hungarian Gigaword Corpus; \citealp{oravecz2014hungarian}).
This means there should not be a frequency bias affecting the realisation of \textsc{IS}.
Admittedly, characters are not nearly this well balanced in terms of gender: out of 24 characters, only 4 are female on the images, and -- while Hungarian has no grammatical gender -- these are all lexicalised as female words.
This imbalance means we cannot investigate any potential effect of gender on the propensity of topicalisation.
(See Appendix~\ref{app:frequency_counts} for the details of the frequency and gender analysis.)

\begin{table}[]
\centering
\small
\begin{tabular}{lll}
\toprule
\textbf{Word order} & \textbf{Object-focus} & \textbf{Subject-focus} \\
\midrule
SVIO & \#          & preVF      \\
OSVI & \#          & Top-preVF  \\
OIVS & \#          & Top-postVF \\
SIVO & Top-postVF  & neutral               \\
SOVI & Top-preVF   & \#          \\
OVIS & preVF     & \#          \\
\bottomrule
\end{tabular}
\caption{Hungarian word orders and their information-structural function under the two focus conditions.
Orders that are grammatical but infelicitous given
the focus condition are marked with `\#'.
\textbf{S} = subject, \textbf{V} = verb, \textbf{I} = verb modifier (\textit{igekötő}), \textbf{O} = object;
 \textbf{Top-} = topicalisation, \textbf{preVF} = preverbal Focus, \textbf{postVF} = post-verbal Focus.
See Table~\ref{tab:focus_types_full} in Appendix~\ref{app:table} for example sentences and per-NP role tagging.}
\label{tab:focus_types}
\end{table}

\begin{figure*}
    \centering
    \includegraphics[width=\linewidth]{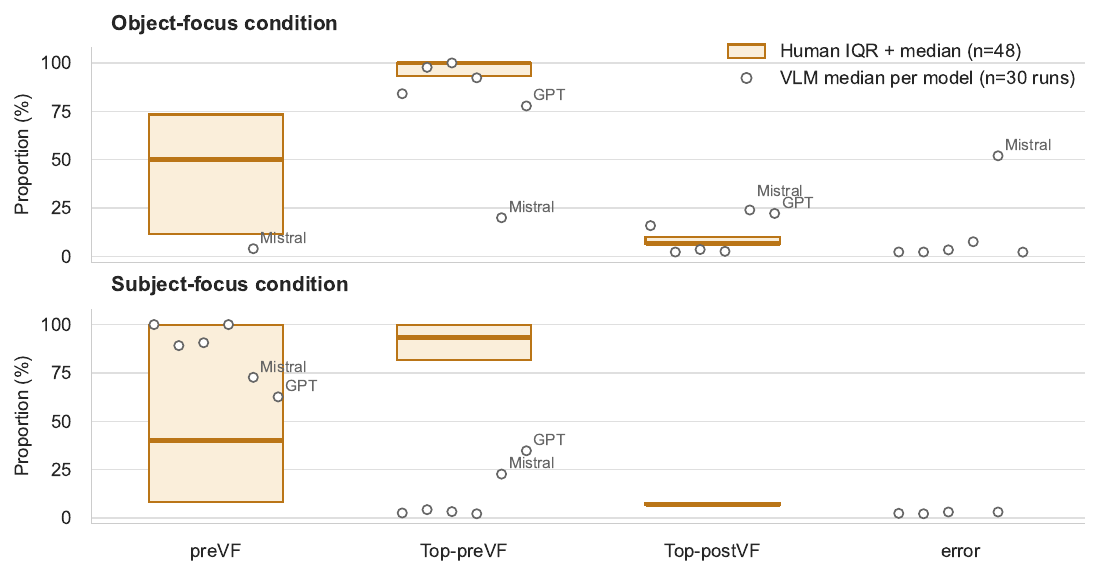}
    \caption{
    Distribution of \textsc{IS}-type proportions in the object-focus (top) and subject-focus (bottom) conditions, including the \textsc{IS}-incongruent \textit{error} condition (\textbf{Top-} = topicalisation, \textbf{preVF} = preverbal Focus, \textbf{postVF} = post-verbal Focus).
    Amber rectangles show the inter-quartile range of per-participant proportions across human participants ($n=48$) with darker horizontal lines at the median.
    The circles show the median proportion across all runs for each of the six VLMs.
    Notable outliers (Mistral Small 3.1 and GPT-4.1) are labelled.
    VLM medians fall within the human range for the object-focus Top-preVF, but sit below it for the subject-focus Top-preVF.
    }
    \label{fig:distributions}
\end{figure*}

\paragraph{Experiments with human participants}

Participants were first shown a prompt requesting terse, short, and complete responses mentioning both characters and the action.\footnote{Prompts can be found in Appendix~\ref{app:prompts}.}
To simulate a real communicative situation and ensure ecological validity of the experiment, we told them their responses help disambiguate between images seen by the person conducting the experiment.
The experiment was run with 51 participants, native Hungarian speakers (1 male, 50 female; mean age 21.18 years, $SD=4.41$) recruited from the authors' academic network.
Each participant gave 30 responses (15 per condition), which allowed generalisation across the variability of human outputs.
Three participants were excluded from the main analyses as their response patterns formed a distinct outlier cluster (Top-postVF-dominant, see Appendix \ref{app:cluster}), leaving $n=48$ for the analyses reported in §\ref{sec:results}.\footnote{%
Due to logging errors, 1.8\% of human responses are missing: thus the total count is 1413 instead of the expected 1440 (48 $\times$ 30).}
Human responses are manually annotated by two Hungarian-speaking expert annotators.
The annotators first extract the sentence type, \textit{i.e.,} the order of the verb (\textbf{V}), the subject (\textbf{S}), the object (\textbf{O}), and the verb modifier (\textbf{I}).
We then follow Table \ref{tab:focus_types} to convert sentence type into \textsc{IS}-types depending on whether the response falls under the subject-focus or object-focus condition.
While sentence types are general, \textsc{IS}-types depend on the question.
Top-preVF and Top-postVF both stand for responses with Topic NPs, differing in whether the Focus NP is preverbal (preVF) or post-verbal (postVF).
In preVF responses, there is preverbal Focus but no Topic, while the default sentences represent the `canonical' word order: the most common word order in Hungarian when the discourse does not require the identification or contrast expressed by the Focus.

\paragraph{Adapting the procedure to VLMs}

The stimuli, the conditions, and the way sentence types are mapped to \textsc{IS}-types carry over unchanged between the human and VLM experiments.
There are differences however.
In contrast to humans, VLMs receive an individual system prompt, the image, and the question in separate queries.
While humans required a more naturalistic setup to ensure the experiments have high ecological validity, we do not need to maintain the pretence that VLMs help an interlocutor disambiguate between images.
VLMs are already post-trained to be helpful communicators.

\paragraph{Model selection}

\begin{table}[]
    \centering
    \resizebox{\columnwidth}{!}{
    \begin{tabular}{llrr}
    \toprule
    Model name & Identifier & Size & \# Runs \\
    \midrule
    Claude Opus 4 & \texttt{anthropic/claude-opus-4} & --- & 30 \\
    Gemini 2.5 Pro & \texttt{google/gemini-2.5-pro-preview} & --- & 31 \\
    GPT-4.1 & \texttt{openai/gpt-4.1} & --- & 30 \\
    \addlinespace[0.4em]
    Mistral Small 3.1 & \makecell[l]{\texttt{mistralai/} \\ \texttt{Mistral-Small-3.1-24B-Instruct-2503}} & 24B & 30 \\
    Gemma 3 12b & \texttt{google/gemma-3-12b-it} & 12B & 30 \\
    Gemma 3 27b & \texttt{google/gemma-3-27b-it} & 27B & 30 \\
    \bottomrule
    \end{tabular}
    }
    \caption{
    Runs per model included in our analyses.
    Model sizes are reported only for non-proprietary models.
    Models were last accessed on January 6, 2026.
    }
    \label{tab:model_runs}
\end{table}

We require models to support both visual input and Hungarian (which surfaces \textsc{IS} in ways English would not), and be instruction-tuned; thus outputs reflect pretraining rather than task-specific fine-tuning.
We also include both proprietary and open-weight models for comparison.
The joint requirement for vision models that support Hungarian substantially limits the eligible pool, especially at smaller model scales (Table \ref{tab:model_runs}).
Since VLM documentation rarely covers linguistic capability, and fluency in medium-resource languages like Hungarian is not given \citep{joshi-etal-2020-state}, two expert native-Hungarian annotators rated 557 responses (91--94 per model) on linguistic metrics using a 5-step Likert scale.
Models score above 4 on most metrics, except Mistral Small 3.1 on naturalness.\footnote{See annotation guidelines and the full boxplot in Appendix \ref{app:gramm}.}

\paragraph{Sampling and execution}

A power analysis demonstrates that even a single run of 47 image $\times$ 2 conditions is more than enough to observe statistically significant differences between the word orders generated for the subject-focus and object-focus conditions.
We also control for the stochasticity of VLMs that may cause their responses to differ from random seed to random seed.
We use a bootstrap convergence procedure in the spirit of Monte-Carlo simulation studies \citep{efron1993introduction,morris2019using}.
We find that a maximum of 4 runs already shows a sufficient stability for our results: the 30 runs per VLM\footnote{We carry out 31 runs in the case of Gemini 2.5 Pro.} we carried out is ample (see details on both analysis in Appendix~\ref{sec:app_power}).

We run Gemma 3 12b and 27b, and Mistral Small 3.1 on an Nvidia A40, and use OpenRouter\footnote{\url{https://openrouter.ai}} for GPT-4.1, Gemini 2.5 Pro, and Claude Opus 4.
A single model run represents all queries under the same random seed.

\paragraph{Automated coding of VLM responses}

Given the considerably larger number of VLM responses, manual annotation is replaced with automated parsing.
We use \texttt{huspaCy} v0.9.0 \citep{HuSpaCy:2021,HuSpaCy:2023} with the \texttt{hu\_core\_news\_lg} model (version \texttt{3.8.1}) to recover the same sentence-type representation used in the human pipeline (see Table \ref{tab:focus_types}).\footnote{\url{https://huggingface.co/huspacy/hu_core_news_lg}}
We extract the main verb (\texttt{ROOT}; \textbf{V}) with its corresponding nominal subject (\texttt{nsubj}; \textbf{S}) and direct object (\texttt{obj}; \textbf{O}), as well as its verb modifier (\texttt{compound:preverb}; \textbf{I}).
We refine the parser outputs using several construction-specific heuristics, including restricting analysis to direct children of the \texttt{ROOT} and recovering verb modifer that the parser fails to identify (see Appendix~\ref{app:heuristics}).

Evaluated against a sample of 658 VLM responses annotated by the same expert annotators, our parser pipeline achieves 90.7\% accuracy: in the rest, cc. 7\% represents a mismatch between what the parser and the annotators deem as non-categorisable, and in only 1.5\% of the case there is a parser-annotator disagreement in categorisation.
On the full data, $\sim$15\% of the outputs are non-categorisable, two thirds of which are genuine issues of VLM production.
More than half, 54\% of these non-categorisable responses contain the right verb from the question but no verb modifier, making the Focus unobservable, while an additional 13\% does not even have the right verb.
It is only the remaining one third -- \textit{i.e.}, only about 5\% of all responses -- that can be attributed to genuine parser limitations.
These non-categorisable outputs are excluded from subsequent analyses.

The mapping from sentence type to \textsc{IS}-type then follows Table \ref{tab:focus_types}, exactly as in the human pipeline.
Unlike humans, VLMs do produce responses that conflict with \textsc{IS}, which are marked as errors in the analysis.

\section{Results}\label{sec:results}

\paragraph{VLMs reliably encode \textsc{IS} but produce more formulaic outputs}

Figure~\ref{fig:distributions} shows the proportion of each \textsc{IS}-types produced by humans and by each VLM, split by condition.\footnote{The \textit{default} sentence type is omitted from the figure as it accounts for only 1.9\% of VLM outputs and 0.1\% of human outputs. See Table~\ref{tab:full_breakdown} in Appendix~\ref{app:results} for the full breakdown.}
Across both conditions, \textsc{IS}-congruent categories account for the majority of outputs, indicating that VLMs engage with \textsc{IS} in our experiment.
We report VLM medians without per-run distributions, as run-level variation within each model is small relative to between-model differences (see Appendix \ref{app:figures}).
The shape of this engagement, however, differs by condition.
In the object-focus condition, VLM medians fall within or close to the human inter-quartile range for Top-preVF, the dominant human response.
In the subject-focus condition, VLM medians cluster near zero for Top-preVF, concentrating on preVF instead, sitting well above the human upper quartile.
Two models stand out.
Mistral Small 3.1 produces \textsc{IS}-incongruent \textit{error} responses in roughly half of its object-focus outputs, and GPT-4.1 retains more Top-preVF than other models in subject-focus, sitting closer to the human range than other models.

\paragraph{VLMs exaggerate a human dispreference for object-NP topicalisation}

\begin{figure}%
    \centering
    \includegraphics[width=\linewidth]{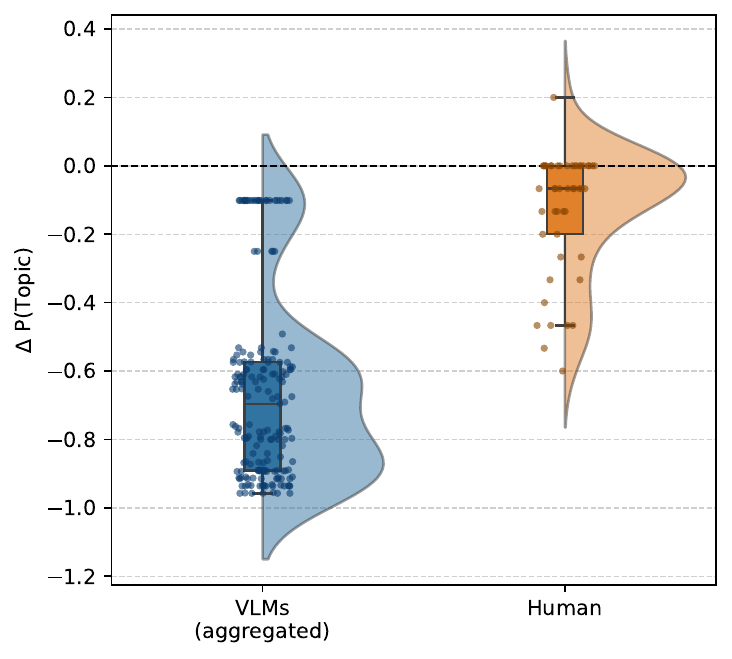}
    \caption{Difference in topicalisation probability in the subject-focus and object-focus conditions, with each point representing one VLM run ($n=30$ per model across six VLMs) or one human participants ($n=48$).
    Humans show graded variation around a small negative shift, VLMs cluster tightly at near-total avoidance of object topicalisation.
    Mistral Small 3.1 is the exception and sits closer to human range likely reflecting its higher error rate (see Appendix \ref{app:figures} for per-model breakdown).
    }
    \label{fig:delta_raincloud}
\end{figure}

In each condition, the character named in the question is established as discourse-old and thus licensed as a Topic in the response: the subject NP in the object-focus, and the object NP in the subject-focus.
Figure~\ref{fig:delta_raincloud} shows the $\Delta$ in topicalisation probability between the two conditions, with each datapoint representing an individual VLM run or human participant.
Both humans and VLMs topicalise less in the subject-focus condition, where this would require object fronting.
What differs sharply is the \textit{shape} of this dispreference.
Humans show graded variation: individual participants range from near-zero change to substantial drops, reflecting different strategies for handling the possibility of object Topics.
VLM runs, by contrast, cluster tightly at near-total avoidance, with few runs near the human median.
The exceptions come from Mistral Small 3.1, whose apparent proximity to humans likely reflects its high error rate rather than its \textsc{IS} sensitivity (see §\ref{sec:discussion}).\footnote{For results broken down per model, see Figure \ref{fig:delta_all} in Appendix \ref{app:figures}.}
Aggregated, VLMs drop from 92.5\% topicalisation in the object-focus condition to 10.4\% in the subject-focus condition (an 82.1-point decrease).
Humans drop 13.4 points: from 81.6\% to 68.2\%.
This near-categorical avoidance, rather than the graded dispreference seen in humans, is sign of the over-regularisation pattern we return to in §\ref{sec:discussion}.

\paragraph{Definiteness tracks Focus in humans but grammatical role in VLMs}

\begin{figure}
    \centering
    \includegraphics[width=\linewidth]{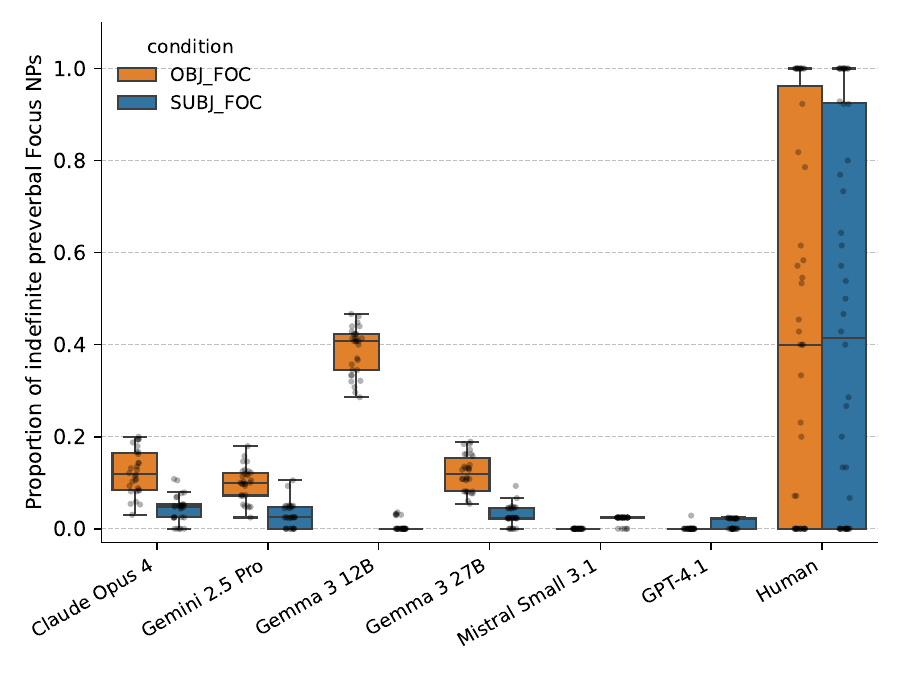}
    \caption{
    Proportion of indefinite NPs in the preverbal Focus position, split by focus condition, per VLM and for human participants. Humans show high indefiniteness in both conditions; most VLMs show low indefiniteness while statistically favouring indefinite objects; Mistral Small 3.1 and GPT-4.1 deviate (see Appendix~\ref{app:stats} for statistical breakdown).
    }
    \label{fig:definiteness}
\end{figure}

Indefinite Focus NPs are expected at higher rates when encoding new information (§\ref{subsec:pressures}).
Figure~\ref{fig:definiteness} shows that humans vary widely on this dimension: some participants produce exclusively indefinite Focus NPs, others exclusively definite, with the rest spread across the range.
VLMs, by contrast, show far less variability, producing almost entirely definite Focus NPs -- with Gemma 3 12B as the only exception (around 40\% indefinite).
A second difference is that VLMs, but not humans, show an asymmetry between conditions.
VLMs produce indefinite Focus NPs more often when the Focus is an object than when it is a subject, an asymmetry that tracks grammatical role rather than discourse status.
For VLMs, an omnibus Kruskal-Wallis test across model-condition groups reveals a significant effect of condition ($H=233.29$, $p<0.0001$).
Bonferroni-corrected Wilcoxon signed-rank tests confirm a significant subject-object asymmetry for four models: Claude Opus 4, Gemini 2.5 Pro, Gemma 3 12B, and Gemma 3 27B (full pairwise tests in Appendix~\ref{app:stats}).
Two models deviate.
Mistral Small 3.1 shows the opposite asymmetry, and GPT-4.1 produces near-categorical definiteness for object Focus NPs in both conditions, with no significant difference between them.
These deviations likely reflect distinct factors, discussed in §\ref{sec:discussion}.

\section{Discussion}\label{sec:discussion}

Two additional pressures (RQ2a, RQ2b) interact with the core \textsc{IS} assignment introduced in §\ref{subsec:is}.
Most directly, Topic and Focus functions are assigned based on what is already established in the discourse and what is newly contributed, and Hungarian word order makes this assignment overt.
Definiteness aligns with the same distinction: Topics referring to established entities tend to be definite, while Foci introducing new referents tend to be indefinite.
A third pressure cuts across these: grammatical role independently favours subjects as Topics and as preceding objects; objects, in turn, tend to be indefinite.
In our experimental setup, these pressures align in the object-focus condition (the discourse-old element is the subject, the unmarked Topic) and conflict in the subject-focus condition: marking the discourse-old object as Topic.
The behaviour we observe -- convergence with humans under alignment, divergence under conflict -- is the central finding of the paper, and is what the rest of this discussion develops.

In the object-focus condition, discourse-driven and grammatical-role-driven pressures align: the discourse-old referent is the subject, which is also the natural Topic.
In the subject-focus condition, they conflict.
Discourse status mandates marking the object as Topic, but grammatical-role priors disfavour this.
Humans navigate this conflict by prioritising discourse status: while individuals diverge in tactics, they broadly maintain similar \textsc{IS}-type distributions (Figure \ref{fig:distributions}) and topicalisation rates (Figure \ref{fig:delta_raincloud}) across the two conditions, and show no subject-object asymmetry in Focus definiteness (Figure \ref{fig:definiteness}).

\paragraph{VLMs behave differently to humans}

While VLMs succeed in filling the Focus position with the new information elicited by the question (shown by the low rate of \textsc{IS}-incongruent \textit{error} responses; Figure \ref{fig:distributions}), their handling of Topic is less human-like.
Under alignment (object-focus), they match humans, overwhelmingly topicalising the subject.
Under conflict (subject-focus), they largely avoid object topicalisation, falling back on preverbal-Focus structures without an overt Topic (Figure \ref{fig:delta_raincloud}).
The same grammatical bias appears in definiteness: VLMs produce more indefinite NPs when the Focus is an object than when it is a subject (Figure \ref{fig:definiteness}).

The convergence of the two patterns -- topicalisation and definiteness -- suggests that VLMs are not failing at \textsc{IS} \textit{per se}, but are latching onto grammatical-role correlates of \textsc{IS} as a substitute.
Therefore, VLM behaviour looks less like flexible discourse-sensitive production and more like the application of high-probability structural templates anchored in grammatical role.
By contrast, human variability reflects the availability of multiple felicitous information-structural options and diverging strategies (see Appendix \ref{app:cluster}), a distribution from which models reproduce only a subset.
This resembles \textit{mode collapse} in generative modelling, where a model represents only a limited subset of the target distribution rather than the full diversity of plausible outputs. 
In this sense, VLMs may be linguistically well-formed while still distributionally deficient.
The responses they generate are fluent and acceptable, but they fail to preserve the range of human strategies expected under the same discourse conditions \citep{kirk2024understanding,zhang2025verbalized}.
Other factors related to the post-training of the models, including instruction tuning \citep{wei2022finetuned} and reinforcement learning from human feedback (RLHF; \citealp{ouyang2022training}) might also impact the diversity of model outputs.

\paragraph{Impact of model capabilities}

Models split differently across measures.
On topicalisation, five of the six VLMs cluster together (Claude Opus 4, Gemini 2.5 Pro, Gemma 3 12B, Gemma 3 27B, and GPT-4.1), with Mistral Small 3.1 as the outlier (Figure \ref{fig:delta_all}).
On \textsc{IS}-type distribution, the picture shifts.
GPT-4.1 patterns closer to Mistral than to the other four (Figure \ref{fig:distributions}).
On definiteness, GPT-4.1 stands alone, showing near-categorical regularisation absent in any other model (Figure \ref{fig:definiteness}).

Two factors likely contribute to these splits.
First, three of the four consistently-clustering models come from Google (Gemini 2.5 Pro, Gemma 3 12B, Gemma 3 27B), with the two Gemma variants sharing architectural lineage and training data.
This makes the convergence of these models unsurprising.
Second, Mistral Small 3.1's behaviour is best read against its Hungarian competence: it ranked lowest on our grammaticality evaluation (§\ref{sec:methods}, Figure \ref{fig:likert}) and has a substantially higher error rate, suggesting insufficient Hungarian skills to engage with \textsc{IS} at all.
GPT-4.1 is the harder case, as it produces fluent, grammatical Hungarian, but it distinctively over-regularises.
This may reflect stronger post-training pressure towards consistent outputs rather than a competence gap.

\paragraph{Implications for evaluation}

Our work has implications for how linguistic competence in VLMs should be evaluated.
A model may produce grammatical and discourse-congruent answers while still failing to capture the variability characteristic of human production. 
For information structure in particular, the target is not a single correct word order, but a distribution of acceptable choices conditioned by discourse status, grammatical role, definiteness, and speaker strategy.
Evaluation frameworks that score against a single reference, or that collapse human responses into a majority answer, will miss this distributional dimension --- and with it, a substantial part of what it means to use language in context.

\section{Conclusion}\label{sec:conclusion}

In this paper, we use Hungarian as a testbed, a language where \textsc{IS} surfaces overtly in word order.
We evaluate six contemporary VLMs, including both proprietary and open-weight models, and compare their outputs to human responses collected under analogous experimental conditions.
Our results show that VLMs are sensitive to \textsc{IS}, producing discourse-congruent responses at higher rates.
However, behaviour also differs from human production systematically.
VLMs offer less variable outputs than humans -- a pattern that resembles mode collapse discussed in generative modelling -- and favour linguistic choices that stem from grammatical role priors.
They strongly favour subject topicalisation, and show a subject-object asymmetry in Focus definiteness absent in human data.
These results suggest that VLMs acquire some discourse-conditioned form-function mappings, but realise them through narrower and more regularised production strategies than human speakers.
These findings argue for VLM evaluation that accounts for the distributional shape of human production, not only the accuracy of content.

\section*{Limitations}

Our research covers a single language, Hungarian, precisely because \textsc{IS} is overt in its word order.
This imposes a limit on the applicability of our findings to languages that express \textsc{IS} in different ways.
Generalisability is further challenged by how we constrain the linguistic space in our experimental setup.
Images and questions cover a set of transitive and semantically reversible events acted out by animate human characters, and actions are expressed with verbs with verb modifiers.
This means we test \textsc{IS} in a comparatively narrow slice of language use instead of the broader variation of discourse phenomena that also exists outside of \textsc{IS}.
Furthermore, while the characters in the images and questions are all referred to with comparably common vocabulary items, thus controlling for frequency effect, there is an imbalance in their apparent gender -- 20 male characters to 4 female ones -- meaning we cannot investigate the potential impact gender might have on topicalisation.

Human variation is approximated by a human sample with a distinct demographic skew: young and mostly female Hungarian speakers.
Whether these patterns of human variability generalise to demographically broader samples is an empirical question we leave to future work.
Model behaviour is approached through a limited set of models, half of which come from the same provider.
This stems from the necessity for models that cover both visual input and show at least basic competence in Hungarian, a medium-resource language.
The requirement for Hungarian competence excluded smaller models.
Even within our selected set, Mistral Small 3.1's behaviour appears to be limited by its Hungarian proficiency.
Furthermore, experiment replication is made difficult by the fact that half of our models are proprietary, and may be silently updated.
We also did not explore prompt variation or temperature settings, both of which could affect the patterns we report.
Our analogy with mode collapse is not supported by further experiments and the lack of diversity of VLM outputs compared to human outputs may just as well emerge from post-training related factors.

Since we do not have access to the training data, we cannot assess whether the observed results come from bias in the training data.
All we can claim is that under the same controlled experimental conditions, VLMs exhibit a substantially stronger subject-Topic preference than human participants.
The human experiment suggests that this preference may reflect a genuine tendency in Hungarian usage.

Extending this evaluation framework to other discourse-configurational languages, broader stimulus sets, and more diverse model families would establish the generality of the cue-conflict/over-regularisation pattern we observe.
Directly comparing VLM attention patterns with human eye-tracking data is out of our scope; we believe, however, that future work could benefit from this analysis.

\section*{Ethical Considerations}

The procedure of human data collection was reviewed and approved by the Scientific and Research Ethics Committee of the Hungarian Health Science Council (ETT TUKEB)\footnote{\url{https://ett.okfo.gov.hu/tukeb/}}, the competent body for human research in Hungary.
Procedures conform to the principles of the ACL Code of Ethics.
Participants were informed of the purpose of the study, the nature of the task, and the recording of their responses; they gave informed consent and could withdraw at any time.
Participation was voluntary and unpaid, reflecting the short duration and low burden of the task.
Responses were stored and processed in anonymised form.
The visual stimuli do not depict identifiable individuals.
The anonymised dataset will be made available upon publication.
Annotation was carried out by two expert annotators: one is a co-author of this paper, and the other was compensated at a rate consistent with local academic pay scales through a research grant (to be named in the Acknowledgements).
We used generative AI (Claude) to assist with visualisation code, post-editing, and parts of the literature review, all manually verified by the authors.

\section*{Acknowledgments}

MF and JB are funded by the Carlsberg Foundation, under the Semper Ardens: Accelerate programme (project no. CF21-0454) and the Novo Nordisk Foundation under the Ascending Data Science Investigator programme (grant no. NNF24OC0092972).
TK's research was supported by the National Research, Development and Innovation Office of Hungary (NKFIH), grant no. 143301.
We thank Réka Vágner for her assistance with annotation.

\bibliography{custom,anthology-1,anthology-2}

\appendix

\section{Full Table}\label{app:table}

See Table \ref{tab:focus_types_full}.

\begin{table*}[ht]
\centering
\setlength{\tabcolsep}{5pt}
\resizebox{\linewidth}{!}{%
\begin{tabular}{llllllll}
\toprule
 & & \multicolumn{3}{c}{\textbf{Object-focus condition}} & \multicolumn{3}{c}{\textbf{Subject-focus condition}} \\
\cmidrule(lr){3-5} \cmidrule(lr){6-8}
\textbf{Type} & \textbf{Example Sentence} & \textbf{Function} & \textbf{Subj.} & \textbf{Obj.} & \textbf{Function} & \textbf{Subj.} & \textbf{Obj.} \\
\midrule

SVIO &
\begin{tabular}[t]{@{}l@{\hspace{8pt}}l@{\hspace{8pt}}l@{\hspace{8pt}}l@{}}
$[_S$A fiú$]$   & $[_V$festi$]$  & $[_I$le$]$  & $[_O$a lányt$]$. \\
$[_S$The boy$]$ & $[_V$paints$]$ & $[_I$VM$]$  & $[_O$the girl$]$.
\end{tabular}
& \# & F & PVF & \textbf{preVF} & F & PV \\
\addlinespace

OSVI &
\begin{tabular}[t]{@{}l@{\hspace{8pt}}l@{\hspace{8pt}}l@{\hspace{8pt}}l@{}}
$[_O$A lányt$]$   & $[_S$a fiú$]$   & $[_V$festi$]$  & $[_I$le$]$ \\
$[_O$The girl$]$  & $[_S$the boy$]$ & $[_V$paints$]$ & $[_I$VM$]$
\end{tabular}
& \# & F & T & \textbf{Top-preVF} & F & T \\
\addlinespace

OIVS &
\begin{tabular}[t]{@{}l@{\hspace{8pt}}l@{\hspace{8pt}}l@{}}
$[_O$A lányt$]$  & $[_{IV}$lefesti$]$   & $[_S$a fiú$]$. \\
$[_O$The girl$]$ & $[_{IV}$VM-paints$]$ & $[_S$the boy$]$.
\end{tabular}
& \# & PV & T & \textbf{Top-postVF} & PVF & T \\
\addlinespace

SIVO &
\begin{tabular}[t]{@{}l@{\hspace{8pt}}l@{\hspace{8pt}}l@{}}
$[_S$A fiú$]$   & $[_{IV}$lefesti$]$   & $[_O$a lányt$]$. \\
$[_S$The boy$]$ & $[_{IV}$VM-paints$]$ & $[_O$the girl$]$.
\end{tabular}
& \textbf{Top-postVF} & T & PVF & neutral & T & PV \\
\addlinespace

SOVI &
\begin{tabular}[t]{@{}l@{\hspace{8pt}}l@{\hspace{8pt}}l@{\hspace{8pt}}l@{}}
$[_S$A fiú$]$   & $[_O$a lányt$]$   & $[_V$festi$]$  & $[_I$le$]$ \\
$[_S$The boy$]$ & $[_O$the girl$]$  & $[_V$paints$]$ & $[_I$VM$]$
\end{tabular}
& \textbf{Top-preVF} & T & F & \# & T & F \\
\addlinespace

OVIS &
\begin{tabular}[t]{@{}l@{\hspace{8pt}}l@{\hspace{8pt}}l@{\hspace{8pt}}l@{}}
$[_O$A lányt$]$   & $[_V$festi$]$  & $[_I$le$]$ & $[_S$a fiú$]$. \\
$[_O$The girl$]$  & $[_V$paints$]$ & $[_I$VM$]$ & $[_S$the boy$]$.
\end{tabular}
& \textbf{preVF} & PV & F & \# & PVF & F \\
\addlinespace

SOIV &
\begin{tabular}[t]{@{}l@{\hspace{8pt}}l@{\hspace{8pt}}l@{}}
$[_S$A fiú$]$   & $[_O$a lányt$]$  & $[_{IV}$lefesti$]$. \\
$[_S$The boy$]$ & $[_O$the girl$]$ & $[_{IV}$VM-paints$]$.
\end{tabular}
& \# & --- & --- & \# & T & --- \\

\bottomrule
\end{tabular}
}
\caption{Hungarian word orders and their information-structural function under the two focus conditions: the \textbf{object-focus condition} (answering \textit{Kit fest le a fiú?} `Who is the boy painting?') and the \textbf{subject-focus condition} (answering \textit{Ki festi le a lányt?} `Who is painting the girl?').
Each example is followed by a constituent-by-constituent English gloss in the original Hungarian order; \textbf{VM} glosses the verb modifier \textit{le}.
Orders that are grammatical but infelicitous given the focus condition are marked with `\#'.
\textbf{S} = subject, \textbf{V} = verb, \textbf{I} = verb modifier (\textit{igekötő}), \textbf{O} = object. \textbf{F} = Focus, \textbf{T} = Topic, \textbf{PV} = post-verbal NP, \textbf{PVF} = post-verbal Focus.}
\label{tab:focus_types_full}
\end{table*}

\section{All Results}\label{app:results}

Results in Table \ref{tab:full_breakdown}.

\begin{table*}[htbp!]
\resizebox{\linewidth}{!}{
\begin{tabular}{llrrrrrrr}
\toprule
 & Sentence Type & default & preVF & Top-preVF & Top-postVF & error & \textit{n} \\
Model Name & Condition &  &  &  &  &  &  \\
\midrule
\multirow[t]{3}{*}{\texttt{anthropic/claude-opus-4}} & OBJ\_FOC & 0.0 & 0.0 & 84.8 & 15.0 & 0.2 & \textit{1298} \\
 & SUBJ\_FOC & 0.2 & 99.7 & 0.1 & 0.0 & 0.1 & \textit{1221} \\
 & Total & 0.1 & 48.3 & 43.7 & 7.7 & 0.1 & \textit{2519} \\
\midrule
\multirow[t]{3}{*}{\texttt{google/gemini-2.5-pro-preview}} & OBJ\_FOC & 0.0 & 0.0 & 96.3 & 2.0 & 1.8 & \textit{1371} \\
 & SUBJ\_FOC & 6.5 & 90.0 & 3.5 & 0.0 & 0.1 & \textit{1404} \\
 & Total & 3.3 & 45.5 & 49.3 & 1.0 & 0.9 & \textit{2775} \\
\midrule
\multirow[t]{3}{*}{\texttt{google/gemma-3-12b-it}} & OBJ\_FOC & 0.0 & 0.0 & 98.3 & 1.0 & 0.7 & \textit{841} \\
 & SUBJ\_FOC & 7.1 & 90.1 & 0.5 & 0.0 & 2.3 & \textit{1005} \\
 & Total & 3.8 & 49.1 & 45.1 & 0.4 & 1.6 & \textit{1846} \\
\midrule
\multirow[t]{3}{*}{\texttt{google/gemma-3-27b-it}} & OBJ\_FOC & 0.0 & 0.0 & 92.3 & 0.4 & 7.3 & \textit{1132} \\
 & SUBJ\_FOC & 0.0 & 99.6 & 0.4 & 0.0 & 0.0 & \textit{1343} \\
 & Total & 0.0 & 54.0 & 42.5 & 0.2 & 3.4 & \textit{2475} \\
\midrule
\multirow[t]{3}{*}{\texttt{mistralai/Mistral-Small-3.1-24B-Instruct-2503}} & OBJ\_FOC & 0.0 & 3.5 & 18.9 & 27.3 & 50.3 & \textit{715} \\
 & SUBJ\_FOC & 6.3 & 73.0 & 20.2 & 0.0 & 0.4 & \textit{1260} \\
 & Total & 4.1 & 47.8 & 19.7 & 9.9 & 18.5 & \textit{1975} \\
\midrule
\multirow[t]{3}{*}{\texttt{openai/gpt-4.1}} & OBJ\_FOC & 0.0 & 0.0 & 76.5 & 23.4 & 0.1 & \textit{1348} \\
 & SUBJ\_FOC & 2.3 & 62.8 & 35.0 & 0.0 & 0.0 & \textit{1356} \\
 & Total & 1.1 & 31.5 & 55.7 & 11.7 & 0.0 & \textit{2704} \\
\midrule
\multirow[t]{3}{*}{All VLM (aggregated)} & OBJ\_FOC & 0.0 & 0.4 & 81.4 & 11.1 & 7.1 & \textit{6705} \\
 & SUBJ\_FOC & 3.6 & 85.6 & 10.4 & 0.0 & 0.4 & \textit{7589} \\
 & Total & 1.9 & 45.6 & 43.7 & 5.2 & 3.5 & \textit{14294} \\
\midrule
\multirow[t]{3}{*}{Human} & OBJ\_FOC & 0.0 & 18.4 & 81.0 & 0.6 & 0.0 & \textit{705} \\
 & SUBJ\_FOC & 0.1 & 31.6 & 68.1 & 0.1 & 0.0 & \textit{708} \\
 & Total & 0.1 & 25.1 & 74.5 & 0.4 & 0.0 & \textit{1413} \\
\bottomrule
\end{tabular}
}
\caption{Full sentence-type distribution (\%) per model and focus condition. \textit{n} = total sentence count. Top-preVF and Top-postVF are topicalised orders; preVF and default are non-topicalised; error denotes outputs outside the annotation scheme.}
\label{tab:full_breakdown}
\end{table*}

\section{Grammaticality Judgements}\label{app:gramm}

\begin{table*}[ht]
\centering
\renewcommand{\arraystretch}{1.4}
\begin{tabularx}{\textwidth}{@{} l l X X @{}}
\toprule
\textbf{Category} & \textbf{Variable} & \textbf{Description} & \textbf{Scale} \\
\midrule
Lexical appropriateness
    & \texttt{lex\_likert}
    & Are the individual lexical items appropriate choices for referring to the characters and the action depicted in the image? \textit{(Is a clown called a clown?)}
    & 1 = completely inappropriate word choice; 5 = perfect word choice \\
\makecell[tl]{Morphosyntactic\\well-formedness}
    & \texttt{m\_synt\_likert}
    & To what degree does the response conform to the morphological and syntactic rules of Hungarian?
    & 1 = completely ungrammatical; 4 = minor slip; 5 = fully grammatical \\
Semantic well-formedness
    & \texttt{sem\_likert}
    & Independent of form, how well does the meaning of the response correspond to the characters and the event depicted in the image?
    & 1 = semantically anomalous or unrelated to the scene; 5 = fully coherent and faithful to the depicted scene \\
\midrule
Naturalness
    & \texttt{naturalness}
    & Holistic, deliberately subjective impression: how natural does the response sound as an answer to the given question about the given image? Not based on any single linguistic level but on overall impression.
    & 1 = unnatural; 5 = completely natural \\
\bottomrule
\end{tabularx}
\caption{Annotation variables and coding instructions. The first three categories target distinct linguistic levels; \texttt{naturalness} is a holistic judgment on top of these.}
\label{tab:annotation-scheme}
\end{table*}

\begin{figure*}
    \centering
    \includegraphics[width=\linewidth]{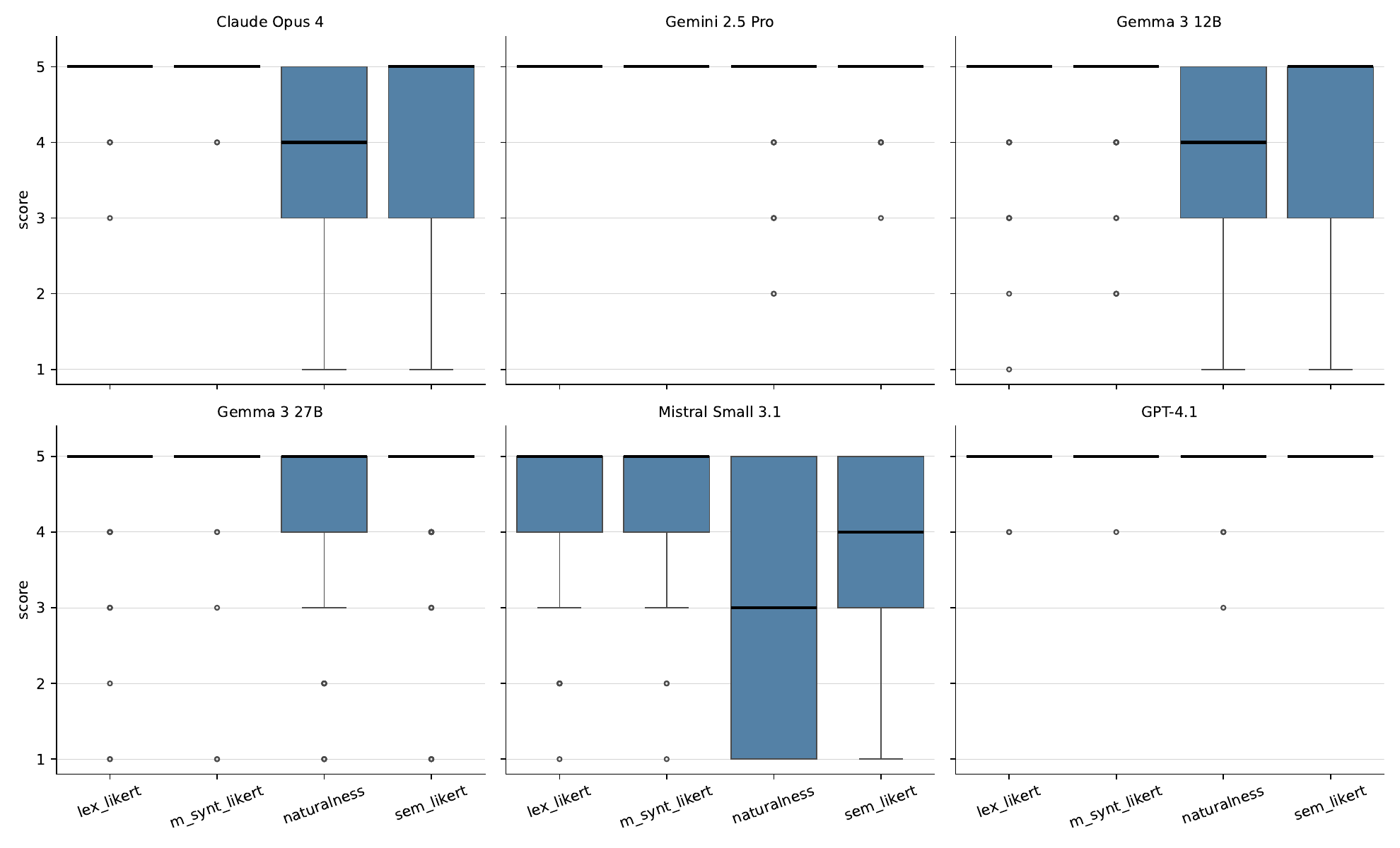}
    \caption{
    The distribution of the Likert scale values for lexical (lex\_likert), and morphosyntactic (m\_synt\_likert) correctness, naturalness, and semantic correctness (sem\_likert).
    }
    \label{fig:likert}
\end{figure*}

The linguistic skills of LLMs and VLMs are strongest in English and other high-resource languages \citep[\textit{inter alia}]{joshi-etal-2020-state}.
Hungarian is a medium-resource language, and VLM fluency is less certain.
We thus evaluate VLM language skills on a set of linguistic metrics on a total of 557 samples (91-94 samples per language model) using two expert annotators who are native speakers of Hungarian, including one of the co-authors of this paper.
The annotators rate the responses in the sample on 5-scale Likert scales reflecting the linguistic categories included in Table \ref{tab:annotation-scheme}.
Figure \ref{fig:likert} showcases the distribution of the Likert scales across the VLMs.

\section{Power Analysis and Bootstrap Convergence}
\label{sec:app_power}

\paragraph{Required Sample Size}

We estimate the required sample size -- \textit{i.e.}, number of responses -- based on 3 runs of GPT-4.1 and Gemma 12b, and 5 runs of Mistral Small 3.1.
These three models are representative of the full set in Table \ref{tab:model_runs}, spanning open-source and proprietary models and a range of Hungarian proficiency.
First, we derive a contingency table across the two conditions of subject-focused and object-focused questions.
We then approximate the effect size using Cramér's $V$ \cite{cramer1946mathematical} using the following formula (\ref{eq:formula}), simplified in our 2-row case representing the two conditions:

\begin{equation}\label{eq:formula}
    V = \sqrt{\frac{\chi^2}{n\cdot(k-1)}} \quad ,
\end{equation}

\noindent
where $\chi^2$ is the result of the chi-square test carried out on the contingency tables and $n$ is the total number of observations in the contingency table.
Finally, $k=2$ for the smallest dimensionality of the contingency table: the two rows representing the two conditions.
We plug $V$ into a goodness-of-fit solver from the \texttt{statsmodels} Python library\footnote{\url{https://www.statsmodels.org/dev/generated/statsmodels.stats.power.GofChisquarePower.html}} with $\alpha=0.05$ and a required statistical power $\pi=0.8$.

This yields that in order to achieve $\pi=0.8$, we need 11 responses for GPT 4.1, 13 for Gemma 12b, and 55 for Mistral Small 3.1; these are numbers well under the 94 --~47 images $\times$ 2 conditions~-- per random seed.
The high required $N$ for Mistral Small 3.1 ($N=55$) likely reflects its limited Hungarian proficiency rather than a weak underlying effect: near-random responding produces a small $V$, inflating the sample size estimate.

\paragraph{Bootstrap Convergence}

We use a boostrap convergence procedure in the spirit of Monte-Carlo simulation studies \citep{efron1993introduction,morris2019using} to compensate for VLM stochasticity.
VLM outputs are non-deterministic, which means repeated runs are needed for reliable distributional estimates.
This analysis determines how many runs are needed for those estimates to stabilise.
For each candidate run $R=1,2,\ldots$, the procedure takes the following steps:

\begin{enumerate}[noitemsep]
    \item Draw $R$ runs at random for every item (\textit{i.e.}, response) and pool these into a condition $\times$ word order contingency table.
    \item Compute Cramér's V described above to estimate effect from the resulting contingency table.
    \item Repeat steps 1 and 2 for 300 bootstrap resamples, record the distribution of Cramér's V at that run count.
\end{enumerate}

\begin{figure*}
    \centering
    \includegraphics[width=\linewidth]{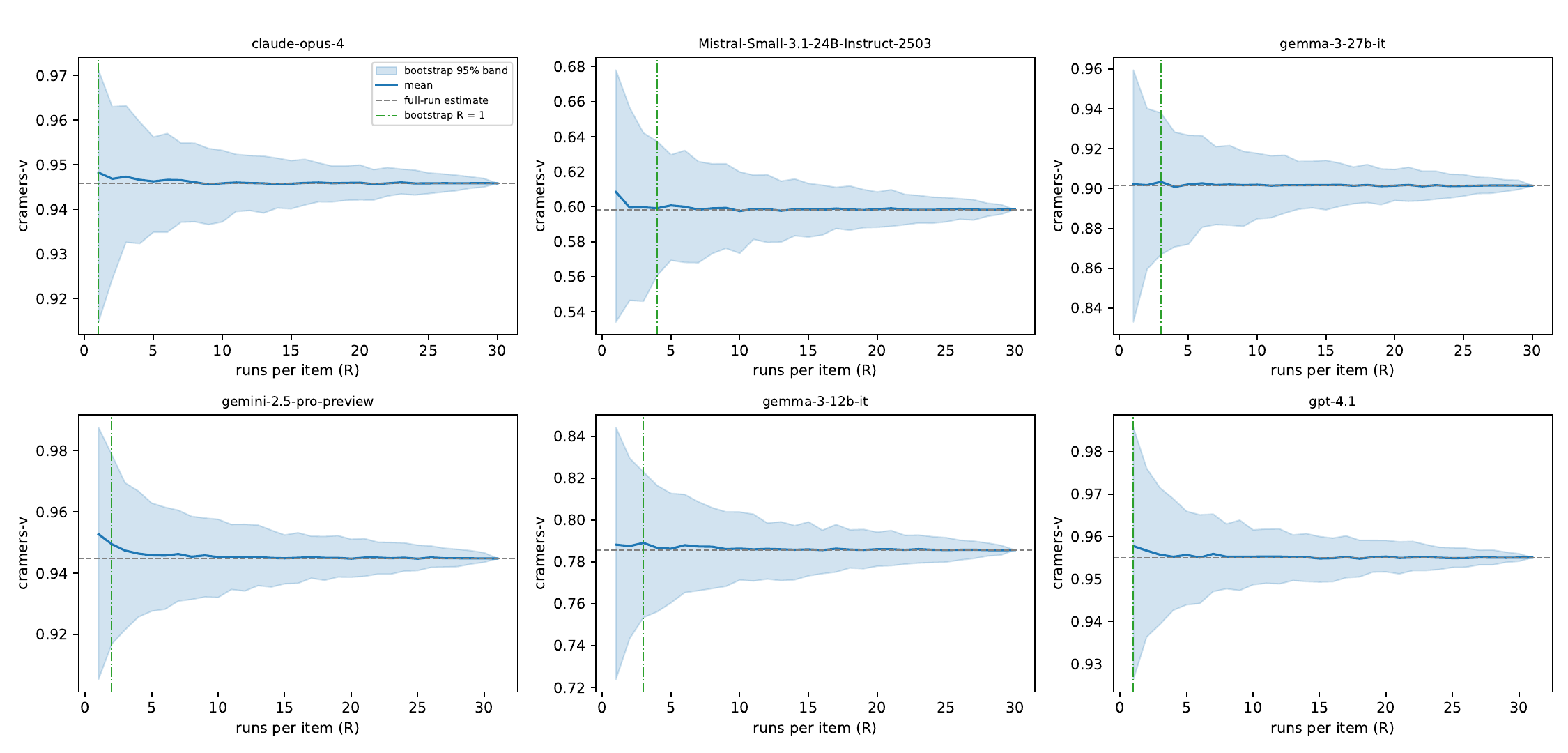}
    \caption{Bootstrap convergence of Cramér's $V$ as a function of runs per item, one panel per model. Blue lines show the bootstrap mean, shaded regions the 95\% percentile band, grey dashed lines the full-sample estimate, and green dash-dotted lines the recommended run count. Note the differing $y$-axis scales across panels.}
\label{fig:run-convergence}

    \label{fig:convergence}
\end{figure*}

For each $R$, we compute the standard deviation of Cramér's V: our stopping criterion is when this falls at or below a small tolerance of 0.02, beyond which collecting more runs would not meaningfully change the estimate.
See Figure~\ref{fig:convergence} for the bootstrap convergence, see Table~\ref{tab:bootstrap-convergence} for the results per VLM.

\begin{table}[t]
\centering
\resizebox{\linewidth}{!}{
\begin{tabular}{lccc}
    \toprule
    \textbf{Model} & \textbf{Runs} & \textbf{Cram\'er's V} & \textbf{Rec.} \\
                   & \textbf{avail.} & \textbf{(full sample)} & \textbf{runs} \\
    \midrule
    Claude Opus 4 & 30 & 0.946 & 1 \\
    GPT-4.1 & 31 & 0.955 & 1 \\
    Gemini 2.5 Pro & 31 & 0.945 & 2 \\
    Gemma 3 27Bb & 30 & 0.902 & 3 \\
    Gemma 3 12b & 30 & 0.786 & 3 \\
    Mistral Small 3.1 & 30 & 0.598 & 4 \\
    \bottomrule
\end{tabular}
}
\caption{Bootstrap-convergence run recommendations. For each model, the recommended run count is the smallest $R$ at which the bootstrap standard deviation of Cramér's V (across 300 resamples) falls at or below the tolerance
of $0.02$. Cramér's V is reported at the full sample.}
\label{tab:bootstrap-convergence}
\end{table}

\section{Experimental Prompts}\label{app:prompts}

Figure~\ref{fig:prompts} show the prompt humans receive prior to the practice round that precedes the experiment, and the system prompt VLMs receive as part of every \texttt{prompt-image-question} triplet in each query: Figure~\ref{fig:prompt-human} and Figure~\ref{fig:prompt-vlm}, respectively.

\begin{figure}[ht]
\centering

\begin{subfigure}{\columnwidth}
\begin{tcolorbox}[
    colback=gray!5,
    colframe=gray!60,
    colbacktitle=gray!40,
    coltitle=black,
    arc=2pt,
    boxrule=0.5pt,
    left=8pt, right=8pt, top=6pt, bottom=6pt,
    title=Practice-phase instructions (spoken to participants),
    fonttitle=\bfseries\small,
    fontupper=\small,
]
\textbf{Hungarian (original):} Most a gyakorlás következik. Kérdéseket fogsz hallani. Minden kérdés után fogsz látni egy képet. Az lesz a feladatod, hogy a kép alapján válaszolj a kérdésre! A választ úgy fogalmazd meg, hogy minden szereplőt megemlítesz! Említsd tehát a szereplőket, és azt a cselekvést, vagy eseményt, amelyben részt vesznek! Kerek, egész mondatban válaszolj, fogalmazz tömören! Hangrögzítő készülékkel a válaszaid rögzítésre kerülnek, ezeket később egy másik személy vissza fogja hallgatni és az általad mondott mondat alapján kell kiválasztania négy kép közül a megfelelőt.

\medskip
\textbf{English translation:} Now the practice phase will follow. You will hear questions. After each question you will see an image. Your task is to answer the question based on the image. Formulate your answer so that you mention every character! That is, mention the characters and the action or event in which they participate! Answer in a complete sentence and be concise! Your answers will be recorded with an audio recording device; these will later be listened to by another person, who will have to choose the appropriate image out of four based on the sentence you produced.
\end{tcolorbox}
\caption{Human prompt.}
\label{fig:prompt-human}
\end{subfigure}

\vspace{6pt}

\begin{subfigure}{\columnwidth}
\begin{tcolorbox}[
    colback=gray!5,
    colframe=gray!60,
    colbacktitle=gray!40,
    coltitle=black,
    arc=2pt,
    boxrule=0.5pt,
    left=8pt, right=8pt, top=6pt, bottom=6pt,
    title=System prompt given to the VLM,
    fonttitle=\bfseries\small,
    fontupper=\small,
]
\textbf{Hungarian (original):} Egy kísérlet résztvevője vagy, válaszolj egy rövid egész mondattal a következő kérdésre a képről, amit adni fogok. A válaszodban említsd meg mindkét szereplőt és a cselekvést, vagy eseményt, amelyet az egyik a másikkal csinál! Fogalmazz tömören!

\medskip
\textbf{English translation:} You are a participant in an experiment; answer the following question about the image I will provide with a short, complete sentence. In your answer, mention both characters and the action or event that one is doing with the other. Be concise.
\end{tcolorbox}
\caption{VLM prompt.}
\label{fig:prompt-vlm}
\end{subfigure}

\caption{Instructions given to (a) human participants during the practice phase, and (b) the vision-language models (VLMs) at inference time.}
\label{fig:prompts}
\end{figure}

\section{Statistical Tests}\label{app:stats}

\paragraph{Statistically significant difference in definiteness between the conditions}
Figure~\ref{fig:definiteness} shows an imbalance in the ratio of indefinite preverbal Focus NPs between the object-focus and subject-focus condition in at least a subset of the VLMs.
In order to confirm the difference is statistically significant, we first carry out a Kruskal-Wallis omnibus test that confirms substantial distributional differences across the model-condition groups ($H=233.29, p < 0.0001$), justifying the per-model follow-ups.
Table \ref{tab:def_stats} shows the per model (and human) result of the Bonferroni-corrected Wilcoxon signed-rank tests carried out between the conditions.

\begin{table*}[ht]
\centering
\small
\renewcommand{\arraystretch}{1.3}
\begin{tabularx}{\textwidth}{@{} l c c l X @{}}
\toprule
\textbf{Model} & \texttt{SUBJ\_FOC} & \texttt{OBJ\_FOC} & \textbf{Direction} & \textbf{Notes} \\
\midrule
Claude Opus 4    & 0.05 & 0.12 & \texttt{OBJ} $>$ \texttt{SUBJ} & Moderate, significant. \\
Gemini 2.5 Pro   & 0.03 & 0.10 & \texttt{OBJ} $>$ \texttt{SUBJ} & Similar magnitude to Claude Opus 4. \\
Gemma 3 12B     & 0.00 & 0.41 & \texttt{OBJ} $>$ \texttt{SUBJ} & Largest effect; \texttt{SUBJ\_FOC} is categorically definite. \\
Gemma 3 27B     & 0.02 & 0.12 & \texttt{OBJ} $>$ \texttt{SUBJ} & Consistent with 12B but attenuated. \\
Mistral Small 3.1 & 0.02 & 0.00 & \texttt{SUBJ} $>$ \texttt{OBJ} & Reversed direction, unique among models. \\
GPT-4.1          & 0.00 & 0.00 & n.s. & Near-categorical definiteness in both conditions. \\
\midrule
Humans           & \multicolumn{2}{c}{$\sim$0.43}  & n.s. & No asymmetry ($p_{\text{bonf}} = 0.834$). High indefiniteness overall is the linguistically expected pattern: preverbal focus in Hungarian introduces new/contrastive information, inviting indefinite marking regardless of grammatical role. \\
\bottomrule
\end{tabularx}
\caption{Median indefiniteness rates by condition for each model and for human participants. The \texttt{SUBJ\_FOC} and \texttt{OBJ\_FOC} columns report the median proportion of indefinite NPs across runs. \textit{Direction} indicates which condition shows higher indefiniteness; \textit{n.s.} = not significant after Bonferroni correction.}
\label{tab:def_stats}
\end{table*}

We used non-parametric tests because the data are bounded proportions from small groups and visual inspection confirmed non-normality.

\section{Clustering Human Results}\label{app:cluster}

\begin{figure*}
    \begin{subfigure}{\columnwidth}
        \centering
        \includegraphics[width=\linewidth]{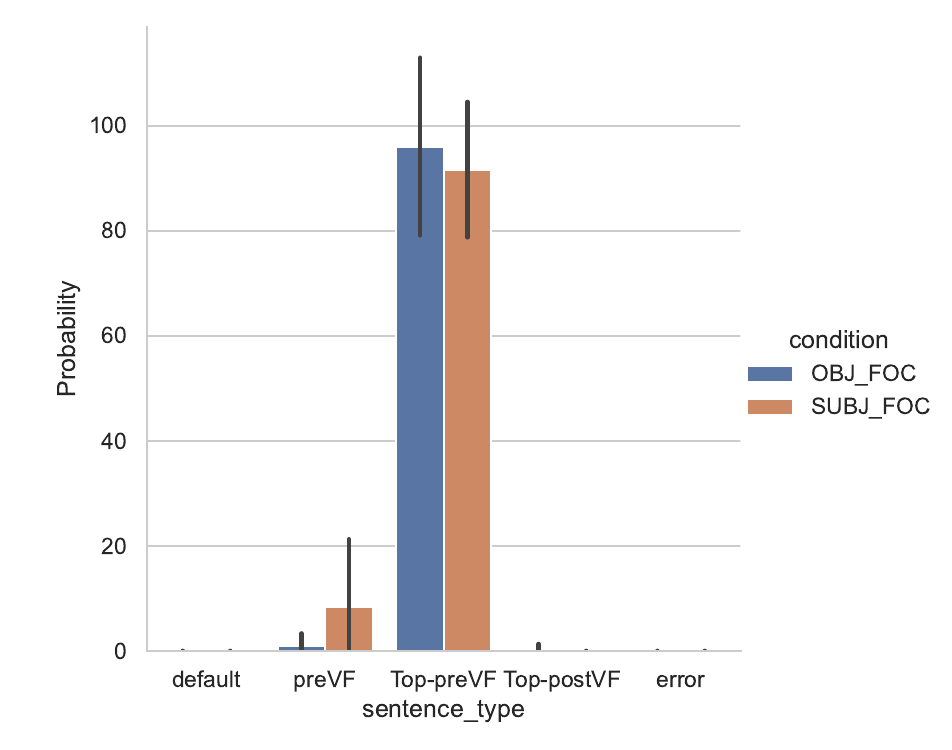}
        \caption{Cluster 1 (35 participants)}
        \label{fig:cluster1}
    \end{subfigure}
    \begin{subfigure}{\columnwidth}
        \centering
        \includegraphics[width=\linewidth]{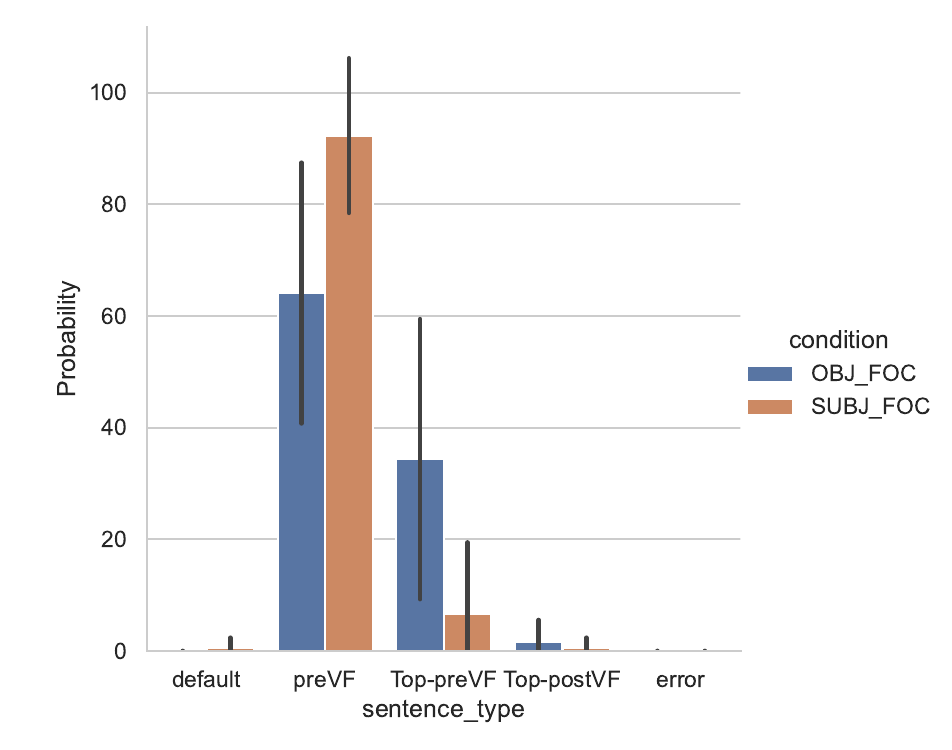}
        \caption{Cluster 2 (13 participants)}
        \label{fig:cluster2}
    \end{subfigure}
    \makebox[\textwidth][c]{%
    \begin{subfigure}{\columnwidth}
        \centering
        \includegraphics[width=\linewidth]{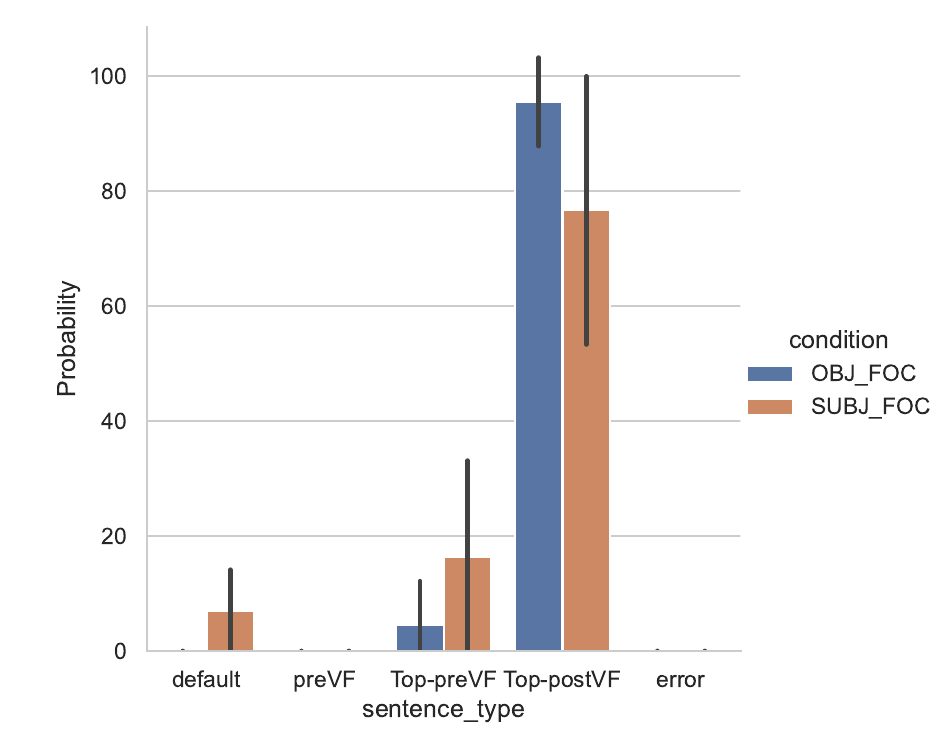}
        \caption{Cluster 3 (3 participants)}
        \label{fig:cluster3}
    \end{subfigure}
    }
    \caption{
    Human clusters created using K-means across subject-focus and object-focus conditions.
    }
    \label{fig:human_clusters}
\end{figure*}

\begin{figure*}
    \centering
    \includegraphics[width=\linewidth]{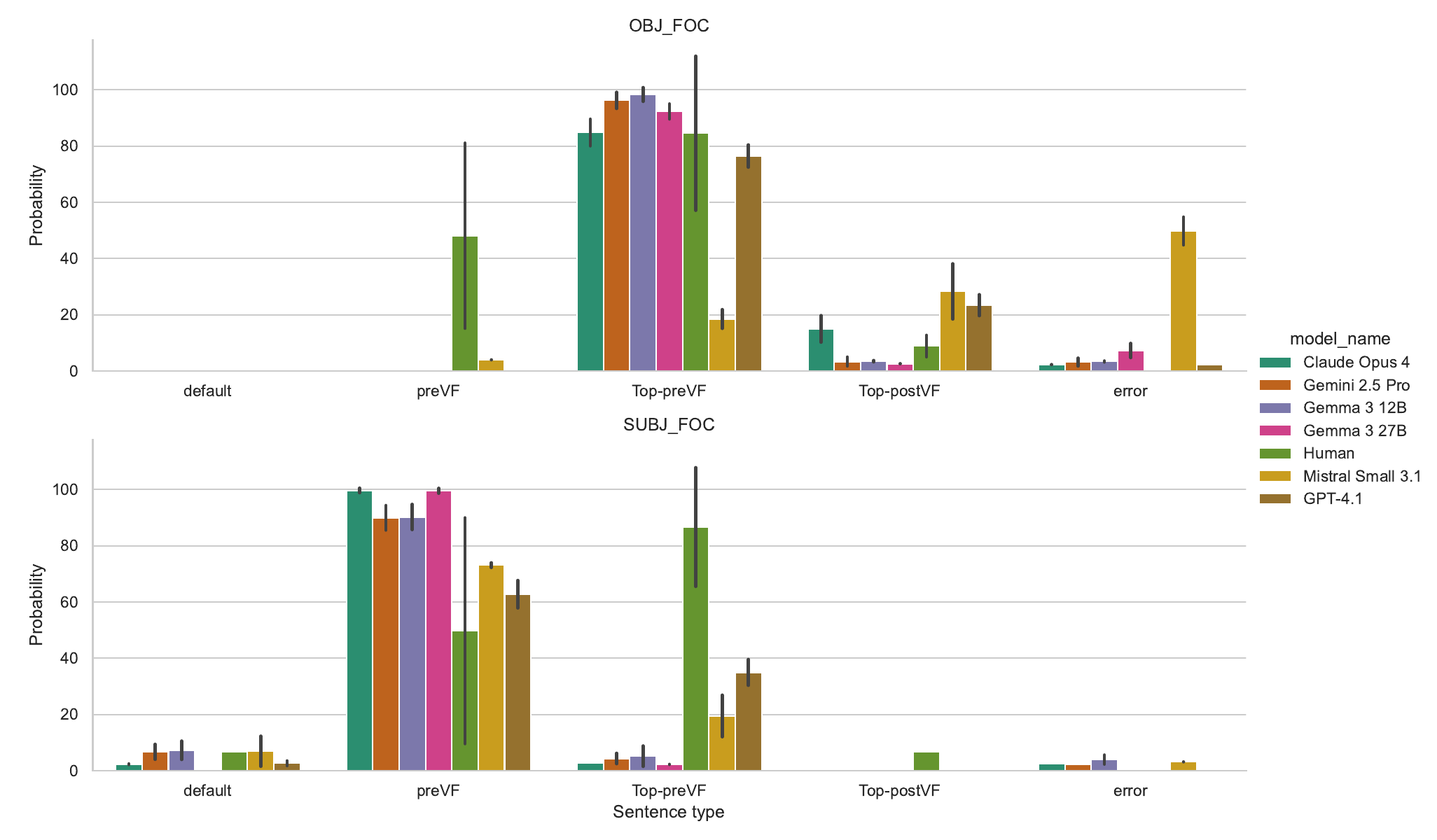}
    \caption{The overall distribution of \textsc{IS}-types across the individual VLMs and humans.
    The error bars reflect variability by representing the standard deviation from the mean.
    }
    \label{fig:sentence_func_bar}
\end{figure*}

\begin{figure*}
    \centering
    \includegraphics[width=\linewidth]{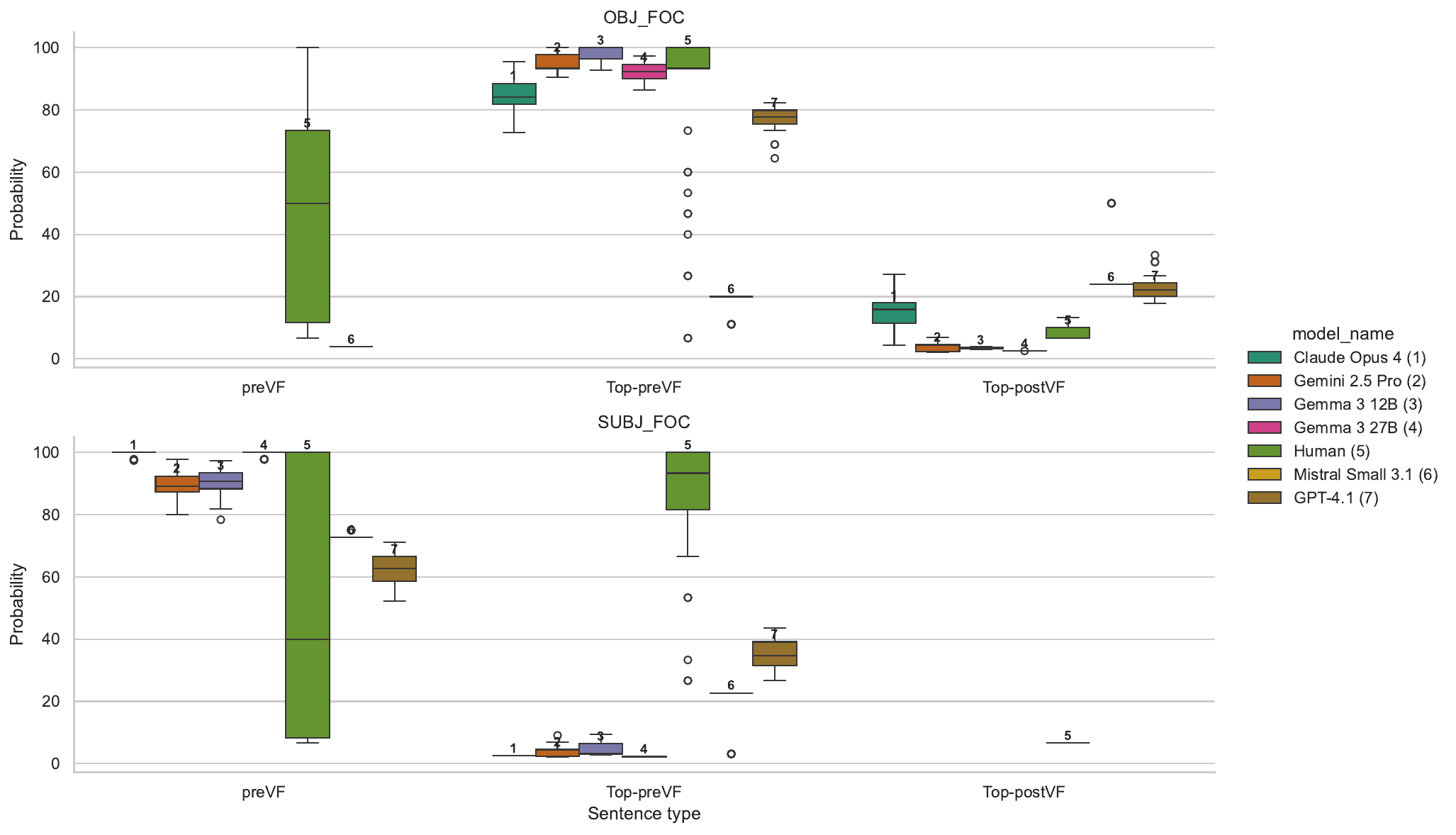}
    \caption{The distribution of \textsc{IS}-types across the individual VLMs and humans.
    }
    \label{fig:sentence_func_box_distr}
\end{figure*}

The error line in Figure \ref{fig:sentence_func_bar} indicates a degree of variability with human participants entirely distinct from VLMs.
When observing the full distribution across all individual runs, Figure \ref{fig:sentence_func_box_distr} indicates that not only are human `runs' are distinctly more variable than the individual VLM runs, variation is considerably higher than across all individual VLMs.
Simply put, different humans are much more variable in their behaviour than the different VLMs we test.
Does this variability reflect an unboundedness of human production, or are there individual patterns in the preference towards topicalisation and the use of focus?

We cluster human productions using the \texttt{scikit-learn} implementation of the K-means clustering algorithm \footnote{\url{https://scikit-learn.org/stable/modules/generated/sklearn.cluster.KMeans.html}}.
First, we calculate the probability of the individual human participants generating a particular sentence type across the subject-focus and object-focus conditions, then fill out the rest of the possible sentence types with 0 values, thus creating $\mathbf{v}_i \in \mathbb{R}^{10}$ for each participant $i$.
Using the elbow method, we identify 3 clusters, and apply the K-means algorithm to get Figure \ref{fig:human_clusters}.

Out of 51 participants, the first cluster represents the majority of 35 (cc. 68.63\%).
They behave more or less symmetrical across the two conditions with a strong preference towards the Top-preVF sentence type.
In some cases, these participants create preVF sentence types, but mainly in the subject-focus condition.
The second cluster of 13 participants (cc. 25.49\%) shows much less topicalisation rates than cluster 1, and with a more pronounced asymmetry too.
The majority of responses, especially in the subject-focus condition, are preVF, while in the object-focus condition there is a slightly higher propensity of Top-preVF productions --- with considerable variability as represented by the longer lines representing the standard deviation.
Cluster 1 but especially cluster 2 mirror the tendency discussed with VLMs in Section \ref{sec:results} for a preference against topicalising the object.

The final cluster has only 3 participants (cc. 5.88\%), with considerably different preferences to the other two clusters.
The Top-postVF sentence type dominates across both conditions, virtually representing the only instances of this sentence type generated by humans.
They also show a slight preference against topicalising the object, but the effect is not too strong.
They also produce comparatively more default sentence types without either a topic or a focus.
In the particular context, this is marked behaviour, and makes these 3 participants outliers: their strategy does not fit the pattern show with the other groups and small size of the group makes it impossible to draw conclusions.

\section{Parser Heuristics}\label{app:heuristics}

We post-process the parser outputs in order to reduce errors via a number of heuristics:

\begin{itemize}
    \item \textbf{Direct children of \texttt{ROOT} only}: We only consider immediate dependents of the main verb of the response to decrease noise from processing irrelevant clauses.
    \item \textbf{Preverb insertion}: When the parser does not identify a split preverb in the response, we examine if the preverb in the question is morphologically attached to the main verb.
    \item \textbf{Consecutive duplicate collapse}: Adjacent identical role labels are deduplicated to filter instances where multiple items in a multi-word constituent receive the same label.
    \item \textbf{Required-roles filter}: Any parses that do not contain all four roles \mbox{(S, O, V, I)} are discarded and labelled as \textit{non-categorisable}.
\end{itemize}

\section{Frequency and Gender of Character Names}\label{app:frequency_counts}

\begin{table}[t]
    \centering
    \resizebox{\columnwidth}{!}{
    \begin{tabular}{lllrr}
    \toprule
    Word & Translation & Assoc. Gender & Raw Freq. & Log Freq. \\
    \midrule
    férfi & `man' & male & 463318 & 5.67 \\
    rendőr & `police officer' & male & 186150 & 5.27 \\
    király & `king' & male & 166738 & 5.22 \\
    orvos & `doctor' & male & 147574 & 5.17 \\
    katona & `soldier' & male & 127040 & 5.10 \\
    kislány & `little girl' & female & 65503 & 4.82 \\
    bácsi & `old man' & male & 43365 & 4.64 \\
    néni & `old woman' & female & 41641 & 4.62 \\
    turista & `tourist' & male & 37655 & 4.58 \\
    tűzoltó & `firefighter' & male & 28443 & 4.45 \\
    kisfiú & `little boy' & male & 25775 & 4.41 \\
    focista & `footballer' & male & 19642 & 4.29 \\
    vadász & `hunter' & male & 12948 & 4.11 \\
    szerelő & `mechanic' & male & 11902 & 4.08 \\
    kalóz & `pirate' & male & 9427 & 3.97 \\
    pincér & `waiter' & male & 9389 & 3.97 \\
    boszorkány & `witch' & female & 7902 & 3.90 \\
    bohóc & `clown' & male & 7048 & 3.85 \\
    indián & `Native American' & male & 6459 & 3.81 \\
    ápolónő & `female nurse' & female & 6134 & 3.79 \\
    varázsló & `wizard' & male & 5909 & 3.77 \\
    betörő & `burglar' & male & 4993 & 3.70 \\
    kertész & `gardener' & male & 3962 & 3.60 \\
    juhász & `shepherd' & male & 3365 & 3.53 \\
    \bottomrule
    \end{tabular}
    }
    \caption{
    The table shows both the associated gender and the frequency count -- broken down into raw and logarithmic frequency -- of the lemmas used to name the characters in the image and question dataset used.
    Associated gender is determined based on the apparent gender of the characters on the images.
    The frequency counts are determined based on the Hungarian Gigaword Corpus \citep{oravecz2014hungarian}.
    The mean logarithmic frequency count is 4.35; the standard deviation is 0.61.
    }
    \label{tab:frequency_gender}
\end{table}

\paragraph{Frequency}
We calculate the relative frequency of the the vocabulary items used for naming the characters on the image stimuli by counting lemmas in the Hungarian Gigaword Corpus \citep{oravecz2014hungarian}.
The most common word is férfi `man' with 463,318 occurrences ($\mathrm{log}_{10}=5.67$), while the least common is juhász `shepherd' with 3,365 occurrences ($\mathrm{log}_{10}=3.53$).
The mean of the log frequencies is 4.35, their standard deviation is merely 0.61.

This implies the frequency distribution is high for all words considered.

\paragraph{Gender}
Hungarian has no grammatical gender, but certain words encode gender lexically: in our dataset, only 4 character names out of 24 encode female gender: kislány `little girl', néni `old woman', boszorkány `witch', and ápolónő `female nurse'.
These characters are also represented as female on the visual stimuli.
The rest of the characters are represented as male on the images, whether they lexically encode male gender (király `king' or bácsi `old man'), they are stereotyped to be male (\textit{e.g.}, rendőr `police officer', orvos `doctor', katona `soldier'), or has no such associated bias, most pertinently turista `tourist'.

This is a clear imbalance in gender distribution, which makes it impossible to investigate the potential impact of gender on \textsc{IS}.

\bigskip
\noindent
See Table~\ref{tab:frequency_gender} for a breakdown of the frequency and gender information for all character names.

\section{Additional Figures}
\label{app:figures}

\begin{figure*}
    \centering
    \includegraphics[width=\linewidth]{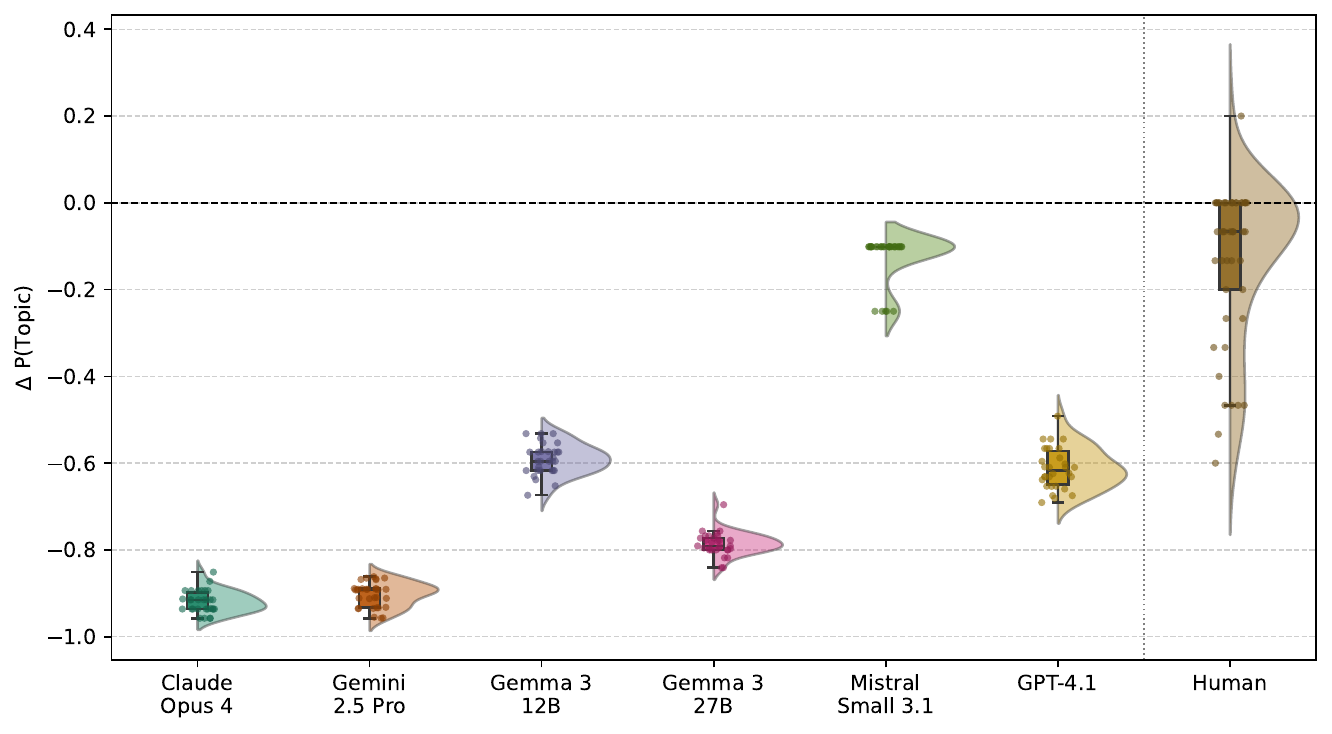}
    \caption{
    Difference in topicalisation probability in the subject-focus condition compared to the object-focus condition between individual runs of VLMs and human participants.
    }
    \label{fig:delta_all}
\end{figure*}

\end{document}